\documentclass[letterpaper, 10 pt, conference]{ieeeconf}  
\IEEEoverridecommandlockouts   
\usepackage{siunitx}
\usepackage{graphics} 
\usepackage{epsfig} 
\usepackage{mathptmx} 
\usepackage{times} 
\usepackage{amsmath} 
\usepackage{mathtools}
\usepackage{amssymb}  
\usepackage{pifont}
\usepackage{subcaption}
\graphicspath{imgs}
\usepackage{xfrac}

\usepackage{graphicx}
\usepackage{booktabs}
\usepackage{multirow}

\newcommand{\myparagraph}[1]{\vspace{4pt}\noindent{\bf #1}}

\title{\LARGE \bf
Learning Temporal Cues by Predicting Objects Move\\for Multi-camera 3D Object Detection}

\author{Seokha Moon, Hongbeen Park, Jungphil Kwon, Jaekoo Lee$^*$, and Jinkyu Kim$^*$
\thanks{S. Moon, H. Park, and J. Kim are with the Department of Computer Science and Engineering, Korea University, Seoul, Republic of Korea.}
\thanks{J. Kwon is with Autonomous Driving Center, Hyundai Motor Company R\&D Division, Seoul, Republic of Korea.}
\thanks{J. Lee is with the College of Computer Science, Kookmin University, Seoul, Republic of Korea}
\thanks{$^*$Co-corresponding authors: J. Lee (jaekoo@kookmin.ac.kr) and J. Kim (jinkyukim@korea.ac.kr)}%
}

\begin{document}

\maketitle
\thispagestyle{empty}
\pagestyle{empty}

\begin{abstract}
In autonomous driving and robotics, there is a growing interest in utilizing short-term historical data to enhance multi-camera 3D object detection, leveraging the continuous and correlated nature of input video streams. Recent work has focused on spatially aligning BEV-based features over timesteps. However, this is often limited as its gain does not scale well with long-term past observations. To address this, we advocate for supervising a model to predict objects' poses given past observations, thus explicitly guiding to learn objects' temporal cues. To this end, we propose a model called DAP (Detection After Prediction), consisting of a two-branch network: (i) a branch responsible for forecasting the current objects' poses given past observations and (ii) another branch that detects objects based on the current and past observations. The features predicting the current objects from branch (i) is fused into branch (ii) to transfer predictive knowledge. We conduct extensive experiments with the large-scale nuScenes datasets, and we observe that utilizing such predictive information significantly improves the overall detection performance. Our model can be used plug-and-play, showing consistent performance gain.

\end{abstract}
\section{INTRODUCTION}
Multi-camera 3D object detection is a crucial task for autonomous vehicles to safely navigate based on understanding their surrounding environment. Recent successes~\cite{bevdet,  bevformer} suggest that each image can be mapped into a frustum of features, rasterizing such frustums into a bird's eye view (BEV) grid. A task-specific object detection head is then applied to detect all objects over the BEV space. Recently, a large performance gain has been obtained by utilizing past BEV features~\cite{bevdet4d, bevstereo, videobev} (i.e., in addition to the current BEV feature, they augment the current and the past BEV features together, thus utilizing temporal cues). However, its gain does not linearly increase with the number of past observations -- its gain is often limited with only two consecutive frames, i.e., the current and previous frames. This may be due to (i) misalignments between BEV features and (ii) suboptimally trained networks that struggle to learn object motions' complicated distribution.

\begin{figure}
    \centering
    \begin{subfigure}[b]{.5\textwidth}
        \includegraphics[width=\textwidth]{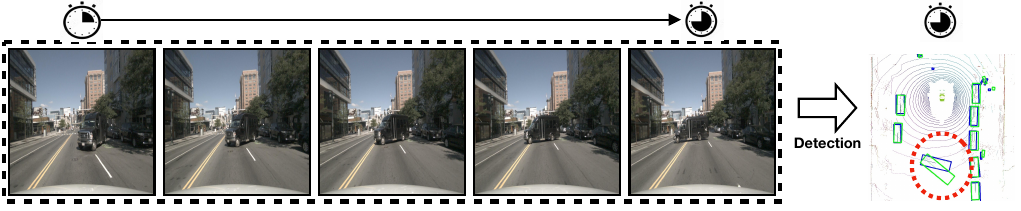}
        \vspace{-1.5em} 
        \caption{Existing Approaches}
        \label{fig:teaser_a}
    \end{subfigure}
    \quad
    \begin{subfigure}[b]{.5\textwidth}
        \includegraphics[width=\textwidth]{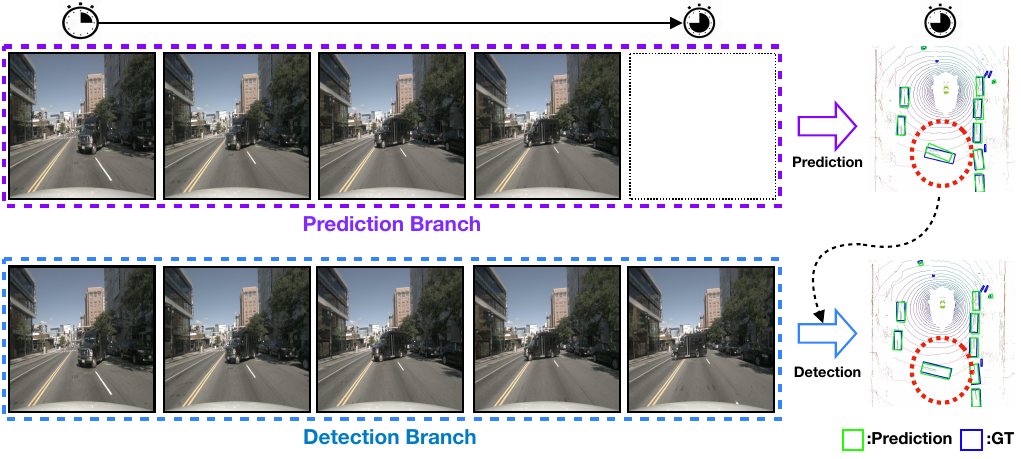}
        \vspace{-1.8em}
        \caption{Our Proposed Approach}
        \label{fig:teaser_b}
    \end{subfigure}
    \caption{Unlike prior multi-camera 3D object detection approaches, which utilize the current and past observations to detect objects, our proposed method regularizes the model by predicting objects' current poses from past observations, and such predictive knowledge is then augmented into object detector, enhancing overall object detection performance.}
    \label{fig:teaser_fig}
    \vspace{-2.0em}
\end{figure}

In this paper, we want to focus on the second issue by regularizing the model to learn better temporal cues. Specifically, we advocate for supervising a model to predict objects' current poses conditioned on past observations only -- a model needs to predict an object's next poses (in the BEV space) based on past observations. We empirically observe that such a regularization and the use of predictive knowledge significantly improves the overall detection performance, potentially owing to better learned temporal cues. See Fig.~\ref{fig:teaser_fig} where we compare ours with existing approaches. 

As shown in Fig.~\ref{fig:main_figure} (a), our model consists of two main modules: (i) Temporal Context Extraction Module and (ii) Context-fused Detection Module. In (i), a network is trained to predict objects' current poses given past BEV-based observations. We further utilize (1) a spatiotemporal BEV encoder to model temporal information effectively and (2) a multi-resolution feature extractor to learn both local (i.e., fine-grained semantic cues per each object) and global (i.e., behavior encoding with large receptive field) representations. In (ii), 3D objects are detected given the current and past observations, where intermediate features from (i) are fused to transfer predictive information. Note that the module (ii) can be easily replaced by conventional BEV-based multi-camera 3D object detection models, such as BEVDet4D~\cite{bevdet4d} and BEVDepth~\cite{bevdepth}. 

To evaluate our proposed approach, we conducted experiments with the publicly available large-scale nuScenes~\cite{nuscenes} dataset. Our model, which is applied to existing BEVDet4D and BEVDepth models, provides a significant improvement, showing comparable performance with the state-of-the-art approaches. Our ablation studies and qualitative analysis further confirm that regularizing a model with predictive information indeed improves the overall detection performance (especially for occluded agents and moving agents), encoding better temporal cues. Note that our code will be publicly available upon publication. 

\section{RELATED WORK}    
\myparagraph{Camera-based Surround View 3D Detection.}
A fundamental approach, as proposed by LSS~\cite{lss}, involves the ``Lift" and ``Splat" methods, which project image features into Bird’s-Eye-View (BEV) representations. Building upon this concept, BEVDet~\cite{bevdet} effectively detects objects using BEV feature. BEVDet4D~\cite{bevdet4d} takes a step further by incorporating information from previous timestamps, demonstrating the valuable contextual information that past data can provide for the present. BEVFormer~\cite{bevformer} introduces a novel method that directly transforms image features into BEV representations using BEV query with BEV shape and deformable attention, drawing inspiration from DETR3D~\cite{detr3d}. BEVDepth~\cite{bevdepth} adopts an approach by directly learning the depth distribution from LiDAR data to construct BEV features. In the context of temporal fusion, VideoBEV~\cite{videobev} effectively reduces computational costs while increasing accuracy by transforming the parallel time fusion method into a recursive fusion method. SOLOFusion~\cite{solofusion} innovates with short-term and long-term fusion modules that leverage historical data for improved environmental understanding. PETRv2~\cite{petrv2} introduces a novel approach to achieve temporal alignment by aligning the 3D coordinates of historical and current frames, facilitating accurate tracking and detection.

\myparagraph{Dense prediction for BEV.}
In the field of autonomous driving, future predictions are typically assessed from the BEV perspective, as the majority of objects and events occur on the ground plane.LSS~\cite{lss} proposes a method of identifying the path of an autonomous vehicle by utilizing BEV features extracted from a camera image. TBP-Former~\cite{tbp-former} proposes a pose-synchronized BEV encoder and spatial-temporal pyramid transformer to accurately map visual features into synchronized BEV space and extract multi-scale features. FIERY~\cite{fiery} uses a conditional variational auto-encoder to generate future instance predictions based on previous BEV features, although it models the entire scene in a single latent code. In contrast, HOPE~\cite{hope} employs latent variables as Gaussian distributions in multi-scale Bird's-Eye-View (BEV) and utilizes aggregators to fuse high-level visual features. It adopts a deep multi-stage encoder-decoder architecture to predict dense occupancy and flow as future motion. UniAD~\cite{uniad} demonstrates that integrating various tasks such as detection, tracking, occupancy, and flow can enhance their individual performance. Furthermore, HoP~\cite{hop} highlights the capability of temporal BEV features to generate BEV features at distinct time intervals.

\section{METHOD}
\begin{figure*}[t]    
    \begin{subfigure}{0.73\textwidth} 
        \centering
        \includegraphics[width=\linewidth]{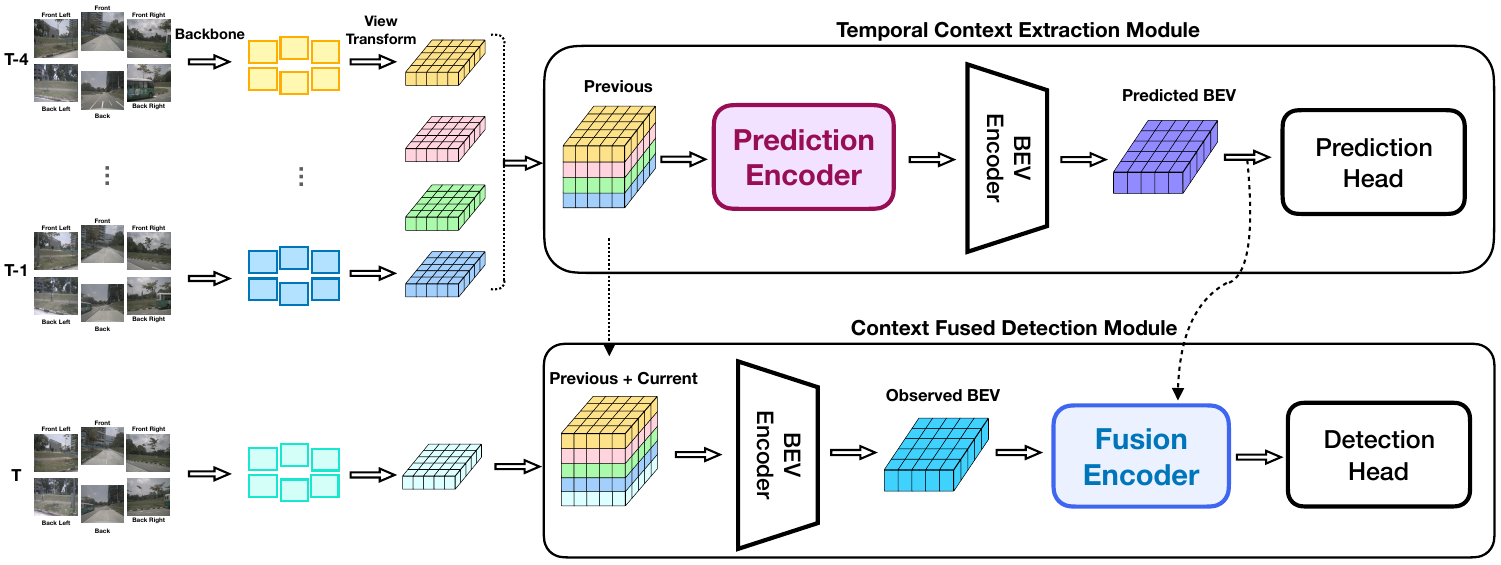}
        \caption{Overall architecture}
        \label{fig:main_a}
    \end{subfigure}
    \begin{subfigure}{0.25\textwidth}
        \begin{subfigure}{\linewidth}
            \centering
            \includegraphics[width=\linewidth]{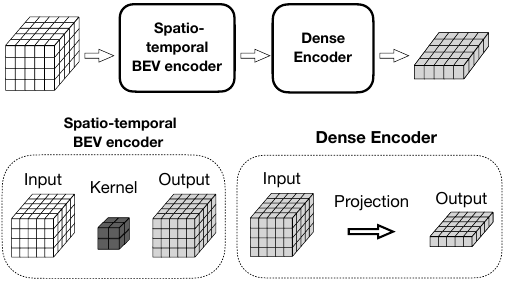}
            \caption{Prediction Encoder}
            \label{fig:main_b}
        \end{subfigure}
        \begin{subfigure}{\linewidth}
            \centering
            \includegraphics[width=\linewidth]{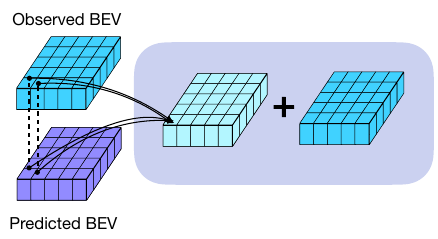}
            \caption{Fusion Encoder}
            \label{fig:main_c}
        \end{subfigure}
    \end{subfigure}
    \caption{Our proposed multi-view 3D object detection architecture. Built upon a conventional BEV(Bird's Eye View)-based multi-view object detection model, our model consists of two main modules: (i) Temporal Context Extraction Module, which predicts objects' current poses conditioned on past BEV-based observations. (ii) Context Fused Detection Module, which detects 3D objects in the scene based on the current and past BEV-based observations. Intermediate BEV feature from the Temporal Context Extraction Module is fused into the Context Fused Detection Module through the Fusion Encoder for the final verdict. }
    \label{fig:main_figure}
    \vspace{-2.0em}
\end{figure*}

\subsection{BEV-based Multi-camera 3D Object Detection}
As shown in Fig.~\ref{fig:main_figure}, our model starts from standard BEV-based multi-camera 3D object detection approaches, which consist of four main modules: (i) a visual encoder that extracts high-level representations given $N$ multi-view images with a backbone network (e.g., ResNet-50), (ii) a view transformer that transforms image-view features onto the BEV space (with a data-driven dense depth predictor) conditioned on extrinsic matrix $E_n\in\mathbb{R}^{3\times 4}$ and intrinsic matrix $I_n\in\mathbb{R}^{3\times 3}$ for $n\in\{1,2,\dots,N\}$, (iii) a BEV encoder that further learns pivotal cues in the BEV space (for better movable objects' scale, orientation, and velocity), and lastly (iv) a task-specific object detection head that detects (movable) objects in the BEV space. Following~\cite{bevdet}, our model is built upon Lift-Splat-Shoot~\cite{lss} for a view transformer, Centerpoint~\cite{centerpoint} for a detection head, and BEVDet~\cite{bevdet} for a visual encoder and a BEV encoder. 

\myparagraph{Learning Temporal Cues with Temporally Aligned BEV Features.}
Recent successes~\cite{bevdet4d,bevformer,videobev,hop} suggest that utilizing temporally augmented BEV features (i.e., fusing the most recent BEV features of previous frames with the current) would boost the overall detection performance by learning temporal cues. Such temporal inputs have been shown to be effective in dealing with objects' speed variations, occluded objects, or depth distribution prediction. We also follow this stream of work~\cite{bevdet4d}, we extract $N$ recent (spatially aligned) BEV features $\mathbf{a}_{n}$ at the current time $t$ for $n\in\{t, t-1, \dots, t-N\}$ from the current and past observations. Similar to \cite{bevdet4d}, we align multiple BEV features according to the current pose of an ego-vehicle, i.e., all previous BEV features are transformed (and interpolated) into the current ego-vehicle's coordinate system based on the measured ego-motion. The aligned BEV features are then concatenated and consumed by a BEV encoder, outputting a further high-level BEV feature $\mathbf{B}^{o}$ as follows: 
\begin{equation}
    \mathbf{B}^{o} =  \texttt{BEVEncoder}_{o} \left(\texttt{Concat}\{\mathbf{a}_{t}, \mathbf{a}_{t-1}, \ldots, \mathbf{a}_{t-N}\}\right)
\end{equation}
where $\texttt{BEVEncoder}_{o}$ represents a BEV encoder. In conventional BEV-based multi-camera 3D object detection approaches, such temporally-augmented feature $\mathbf{B}^{o}$ is then consumed by an object detection head, detecting object attributes, classes, and bounding boxes. 

\subsection{Learning Temporal Cues by Predicting Objects' Current Poses from Past Observations}
Utilizing such a concatenated BEV feature is, however, often limited in improving detection performance for moving objects, where a BEV encoder is often suboptimally trained to leverage the most recent few frames (i.e., the current and previous frames) instead of utilizing whole past observations. Thus, to learn object motions' complicated distributions (for further maximizing temporal information usage), we advocate for supervising a model to predict objects' current poses given past observations and utilizing such intermediate features to improve the overall object detection performance. 

\myparagraph{Spatiotemporal BEV Encoder.}
Similar to the process to obtain a BEV feature $\mathbf{B}^{o}$, we use concatenated (spatially aligned) BEV features of the past observations, i.e., $\{a_{t-1}, a_{t-2}, \dots, a_{t-N}\}$. Note that BEV features are aligned with the ego vehicle's position at each respective time step to be aligned with the current time step. To further capture both temporal and spatial information, we concatenate these aligned features and input them into our Spatiotemporal BEV encoder, denoted as $\mathbb{T}$, which utilizes a 3D ConvNet~\cite{conv3d} to model temporal information effectively. We empirically observe that utilizing such 3D ConvNets preserves the temporal information better than 2D ConvNets (consistent with existing reports~\cite{karpathy2014large}), which often lose their input's temporal signal, under-performing in the objects' future pose prediction task. 
\vspace{-.3em}

\myparagraph{Learning Multi-resolution Features.}
Further, we use a multi-resolution feature learning strategy to learn both local and global representations, i.e., encoding temporal cues for fast-moving vehicles may need larger receptive fields to encode their behaviors (i.e., global), while a model needs to retain fine-grained semantic cues per each object (i.e., local). Thus, we further utilize an additional encoder that maintains high-resolution (or local) representations, coupling the low-resolution (or global) representation in parallel. Specifically, we HRNet~\cite{hrnet}-based encoder that takes outputs from our spatiotemporal BEV encoder as an input, producing multi-resolution feature $\mathbf{B}^{f}$, which (i) will be consumed to predict objects' current poses and (ii) will be augmented into the main branch of object detection. Formally, the multi-resolution feature $\mathbf{B}^{f}$ is defined as follows:
\begin{equation}
    \mathbf{B}^{f} =  \texttt{BEVEncoder}_{p} \left(\mathbb{D}\left(\mathbb{T}\left(\texttt{Concat}\{\mathbf{a}_{t-1}, \mathbf{a}_{t-2}, \ldots ,\mathbf{a}_{t-N}\}\right)\right)\right)
\end{equation}
The term $\text{BEVEncoder}_{p}$ represents the BEVEncoder used in the Temporal Context Extraction Module, while $\mathbb{D}$ denotes the Dense Encoder used in the Prediction Encoder.

\subsection{Fusion Module}
The Predicted BEV feature explicitly predicts the positions of agents within the current BEV frame. This provides insights into where attention should be focused to detect agents at this time point. Therefore, in the Fusion Module, as shown in Fig.~\ref{fig:main_c}, our objective is to effectively integrate the Observed BEV features $\mathbf{B}^{o}$, generated using images from all frames, with the Predicted BEV features $\mathbf{B}^{f}$. This integration aims to enhance our understanding of the current scene.
\begin{multline}
    \label{eq:FDFA}
    FDFA(\mathbf{P}_{i},(\mathbf{B}^{o},\mathbf{B}^{f})) = \\
    \sum_{h=1}^{H}W_{h}\left [\sum_{k=1}^{K}\left(\mathbf{A}_{hik}\bar{W}_{h}\mathbf{B}^{o}_{\left (\mathbf{P}_{i}+\triangle\mathbf{P}_{hik}\right)}+\mathbf{A'}_{hik}\bar{W}_{h}\mathbf{B}^{f}_{\left (\mathbf{P}_{i}+\triangle\mathbf{P}_{hik}\right )}\right) \right]
\end{multline}
\begin{equation}
    \mathbf{\hat{B}}_{\mathbf{P}_{i}} = \mathbf{B}^{o}_{\mathbf{P}_{i}} + FDFA(\mathbf{P}_{i},(\mathbf{B}^{o},\mathbf{B}^{f}))
\end{equation}
To fuse the two features, $\mathbf{B}^{o}$ and $\mathbf{B}^{f}$, each of which has a shape of $\{H, W, C\}$, we use a deformable attention~\cite{deformdetr} based module called Fusion DeFormable Attention($FDFA$). We extract $K$ points for each grid cell from $H$ heads. Given a reference point $\mathbf{P}_{i}$, we ensure that the offset $\triangle\mathbf{P}_{hik}$ in both the $\mathbf{B}^{o}$ and $\mathbf{B}^{f}$ aligns consistently, taking into account the described characteristics. Here, $\triangle\mathbf{P}_{hik}$ represents the offset of the $k$-th point in the $h$-th head with respect to the reference point $\mathbf{P}_{i}$. $\mathbf{B}^{o}{\left(\mathbf{P}_{i}+\triangle\mathbf{P}_{hik}\right)}$ and $\mathbf{B}^{f}{\left(\mathbf{P}_{i}+\triangle\mathbf{P}_{hik}\right)}$ are obtained by bilinear interpolation from their respective features $\mathbf{B}^{o}$ and $\mathbf{B}^{f}$ at the position $\left(\mathbf{P}_{i}+\triangle\mathbf{P}_{hik}\right)$. $\mathbf{A}_{hik}$ and $\mathbf{A'}_{hik}$ represent the attention weights of the feature for the $k$-th point in the $h$-th head with respect to the reference point $\mathbf{P}_{i}$, considering both $\mathbf{B}^{o}$ and $\mathbf{B}^{f}$. Hence, $\sum_{k=1}^{K}\left(\mathbf{A}_{hik}+\mathbf{A'}_{hik} \right ) = 1$. $\bar{W}_{h}$ and $W_{h}$ denote linear layers. $\mathbf{\hat{B}}$ represents the Fusion BEV feature generated by the output of the Fusion Module.

\subsection{Loss Function}
The detection head of the Context Fused Detection Module follows the CenterPoint~\cite{centerpoint} head. It takes the fused BEV feature $\hat{\mathbf{B}}$ as input and predicts several attributes for each agent, including the center heatmap, box scale, velocity, orientation, and the translation from the center heatmap. By minimizing losses for each of these attributes, the model aims to output optimized bounding boxes and their corresponding classes for the detected objects. To achieve this, Gaussian Focal Loss~\cite{cornernet} is employed for the center heatmap prediction, while L1 loss is used for the other attribute predictions. These losses are combined to define the overall loss for the head, denoted as $\mathcal{L}_\text{det}$. Plus, given the past BEV observations $\mathbf{B}^f$, in Temporal Context Extraction Module predicts objects' bounding boxes and their classes in the current time $t$. We minimize the similar loss function $\mathcal{L}_\text{pred}$ together with the main detection loss as follows:
\begin{equation}
     \mathcal{L} =  \lambda_{\text{pred}}\mathcal{L}_{\text{pred}} + \lambda_{\text{det}}\mathcal{L}_{\text{det}}
\end{equation}
where we use hyperparameters $\lambda_{\text{pred}}$ and $\lambda_{\text{det}}$ to control the strength of each loss term. Based on our grid search, we set $\lambda_{\text{pred}}=1$ and $\lambda_{\text{det}}=3$.

\section{EXPERIMENTS}
 \subsection{Setup}

\begin{table}[t]
	\begin{center}
        \vspace{-.5em}
        \caption{Multi-camera 3D object detection performance comparison in terms of NDS (nuScenes Detection Score) and mAP metrics. Our model is applied to existing two approaches, including BEVDet~\cite{bevdet} and BEVDepth~\cite{bevdepth}, and shows improved performance in all metrics. Note that we do not apply CBGS (Cross-balanced Grouping and Sampling) technique, and we use four past frames (i.e., $N=4$). nuScenes~\cite{nuscenes} validation set is used.}
        \label{tab:plugnplay}
        \renewcommand{\arraystretch}{1.2}
        \resizebox{\linewidth}{!}{%
        \begin{tabular}{lccccccc}
        \toprule[.7pt]
            Method  & \textbf{NDS} $\uparrow$ & mAP $\uparrow$  &  mATE $\downarrow$ & mASE $\downarrow$ & mAOE $\downarrow$& mAVE $\downarrow$ \\ \midrule
            BEVDet4D~\cite{bevdet4d} & 0.454 & 0.330 & 0.715 & 0.290 & 0.592 & 0.315 \\
            + Ours & \textbf{0.489} & \textbf{0.374} & \textbf{0.634} & \textbf{0.282} & \textbf{0.575} & \textbf{0.280}     \\ 
            & (3.5\%$\uparrow$) & (4.4\%$\uparrow$) & (8.1\%$\downarrow$) & (0.8\%$\downarrow$) & (1.7\%$\downarrow$) & (3.5\%$\downarrow$)\\ \midrule
            BEVDepth~\cite{bevdepth} & 0.483 & 0.362 & 0.633 & 0.280 & 0.577 & 0.296 \\
            + Ours & \textbf{0.507} & \textbf{0.389} & \textbf{0.575} & \textbf{0.276} & \textbf{0.550} & \textbf{0.271} \\ 
            & (2.4\%$\uparrow$) & (2.7\%$\uparrow$) & (5.8\%$\downarrow$) & (0.4\%$\downarrow$) & (1.7\%$\downarrow$) & (2.5\%$\downarrow$)\\ \bottomrule
        \end{tabular}}
     \end{center}
     \vspace{-2.5em}
\end{table}

\begin{table*}[t]
	\begin{center}
        \caption{3D object detection performance comparison with the recent state-of-the-art approaches. Except for BEVDet~\cite{bevdet}, other approaches utilize temporal information from consecutive multiple frames. Our model is built upon BEVDepth~\cite{bevdepth}, but we would emphasize that our model can be easily applicable to other BEV-based multi-camera 3D object detection methods as well. Also, in this experiment, we use nuScenes~\cite{nuscenes} validation set. Note that all models are trained with CBGS (Class-balanced Grouping and Sampling, \cite{cbgs}) enabled.}
        \label{tab:validation}
        \renewcommand{\arraystretch}{1.2}
        \resizebox{.95\linewidth}{!}{%
        \begin{tabular}{lcccccccccc}
        \toprule[.7pt]
        Method & Backbone & \#Frames & Image Resolution & NDS $\uparrow$ & mAP $\uparrow$  &  mATE $\downarrow$ & mASE $\downarrow$ & mAOE $\downarrow$& mAVE $\downarrow$& mAAE $\downarrow$ \\ \midrule
        BEVDet~\cite{bevdet} & ResNet50 & 1 & 256 x 704 & 0.379 & 0.298 & 0.725 & 0.279 & 0.589 & 0.860 & 0.245 \\
        BEVDet4D~\cite{bevdet4d} & ResNet50 & 2 & 256 x 704 & 0.449 & 0.316 & 0.691 & 0.281 & 0.549 & 0.378 & 0.195 \\
        STS~\cite{sts} & ResNet50 & 2 & 256 x 704 & 0.489 & 0.377 & 0.601 & 0.275 & 0.450 & 0.446 & 0.212 \\
        BEVDet4D~\cite{bevdet4d} & ResNet50 & 8 & 256 x 704 & 0.487 & 0.354 & 0.607 & 0.284 & 0.525 & 0.286 & 0.193 \\
        BEVDepth~\cite{bevdepth} & ResNet50 & 8 & 256 x 704 & 0.519 & 0.399 & 0.571 & 0.281 & 0.463 & 0.278 & 0.206 \\
        BEVStereo~\cite{bevstereo} & ResNet50 & 8 & 256 x 704 & 0.527 & 0.415 & 0.566 & 0.284 & 0.465 & 0.298 & 0.195 \\ 
        SOLOFusion~\cite{solofusion} & ResNet50 & 17 & 256 x 704 & 0.534 & 0.427 & 0.567 & 0.274 & 0.511 & 0.252 & 0.188 \\
        VideoBEV~\cite{videobev} & ResNet50 & 8 & 256 x 704 & 0.535 & 0.422 & 0.564 & 0.276 & 0.440 & 0.286 & 0.198 \\ \midrule
        BEVDepth~\cite{bevdepth} & ResNet50 & 2 & 256 x 704 & 0.484 & 0.362 & 0.617 & 0.274 & 0.480 & 0.393 & 0.203 \\
        BEVDepth~\cite{bevdepth} w/ Ours & ResNet50 & 4 & 256 x 704 & 0.521 & 0.393 & 0.543 & 0.263 & 0.455 & 0.295 & 0.198 \\ 
        BEVDepth~\cite{bevdepth} w/ Ours & ResNet50 & 8 & 256 x 704 & 0.530 & 0.402 & 0.530 & 0.271 & 0.431 & 0.276 & 0.201 \\ 
          & & & & (\textbf{4.6}\%$\uparrow$) & (\textbf{4.0}\%$\uparrow$) & (\textbf{8.7}\%$\downarrow$) & (\textbf{0.3}\%$\downarrow$) & (\textbf{4.9}\%$\downarrow$)  & (\textbf{11.7}\%$\downarrow$) & (\textbf{0.2}\%$\downarrow$) \\\bottomrule
        \end{tabular}}
     \vspace{-1.2em} 
     \end{center}
\end{table*}

\begin{table}[t]
	\begin{center}
        \caption{Categorical comparison of Object Translation, Scale, Orientation and Velocity Estimation Error on nuScenes~\cite{nuscenes} validation set. Both models trained with CBGS~\cite{cbgs}.}
         \label{tab:categorical}
         \renewcommand{\arraystretch}{1.2}
            \resizebox{1.0\linewidth}{!}{%
    	\begin{tabular} {@{}lcccccc@{}} \toprule
            Models  & Agent Type &  mATE $\downarrow$  & mASE $\downarrow$ & mAOE $\downarrow$ & mAVE $\downarrow$  \\ 
            \midrule
            BEVDepth~\cite{bevdepth} &  Vehicles    & 0.447  &  0.175  & 0.125 & 0.266     \\
                              &  Pedestrian    & 0.624  &  0.304  & 0.652  & 0.348    \\  
                              &  Bicycle &   0.466   &  0.276  & \textbf{0.996}  & 0.164       \\ \hline
            Ours              &  Vehicles    &  \textbf{0.435} & \textbf{0.173}   &  \textbf{0.109} & \textbf{0.248}      \\
                              &  Pedestrian    & \textbf{0.556}  &  \textbf{0.295}  & \textbf{0.560}  & \textbf{0.326}      \\
                              &  Bicycle    &  \textbf{0.390} &  \textbf{0.264}  &  1.023 & \textbf{0.133}      \\
            \bottomrule
            \end{tabular}}
     \end{center}\vspace{-2.8em}
\end{table}

\myparagraph{Dataset.}
We conducted experiments on the nuScenes dataset~\cite{nuscenes}, comprising 1000 various scenes from Boston and Singapore. The dataset is divided into 700/150/150 scenes for training, validation, and testing, respectively. Each scene has an approximate duration of 20 seconds, and key samples are annotated at a rate of 2Hz, resulting in a total of 1.4 million object bounding boxes. Captured by 6 cameras to cover the surround view, the images are RGB with a resolution of $900\times 1600$ pixels. Annotations cover 10 classes for object detection, including car, truck, bus, trailer, construction vehicle, pedestrian, motorcycle, bicycle, barrier, and traffic cone. The dataset defines a region of interest within a 51.2-meter radius from the ground plane for 3D object detection.

\myparagraph{Evaluation Metrics.}
We follow the official evaluation protocol of nuScenes~\cite{nuscenes}, which includes five True Positive metrics: Average Translation Error (ATE), Average Scale Error (ASE), Average Orientation Error (AOE), Average Velocity Error (AVE), and Average Attribute Error (AAE). Using a 2m center distance threshold for matching, these metrics comprehensively evaluate system performance in 3D object detection across various dimensions such as transformations, scale, direction, speed, and attribute recognition accuracy. Additionally, we also measure mean Average Precision (mAP) and the NuScenes Detection Score ($\text{NDS} = {1 \over 10}[5~\text{mAP} + \sum_{\text{mTP} \in \mathbb{TP}}{(1 - \textnormal{min}(1, \text{mTP}))}]$ where $\mathbb{TP}$ is five TP metrics.

\myparagraph{Implementation Details.}
For the Image backbone to process images, we employed the ResNet50~\cite{resnet} in conjunction with the Feature Pyramid Network (FPN)~\cite{fpn}. The input images were set to a resolution of $256\times 704$, and we used a dimension $D$ of 256. In addition, experiments were conducted with BEVDet4D~\cite{bevdet4d} and BEVDepth~\cite{bevdepth} as the base. The BEV features, generated through the view transform module, were configured with a shape of $200\times 200$. Each grid cell covered an area of $0.512m \times 0.512m$, with the entire BEV spanning $[-51.2m, 51.2m]$ along the $x$ and $y$ axes. During training, we utilized the AdamW optimizer~\cite{adamw} and trained the model on 8 NVIDIA GeForce RTX 3090 GPUs, with a batch size of 2 samples per GPU. The initial learning rate was set to 2e-4, and we trained for 20 epochs. Subsequently, we continued training for an additional 4 epochs with a reduced learning rate of 2e-5, resulting in a total of 24 epochs.

\subsection{Quantitative Analysis}
\myparagraph{Effect of Leveraging Predictive Information.}
We start by experimenting with existing approaches, including BEVDet4D~\cite{bevdet4d} and BEVDepth~\cite{bevdepth}, to measure the effect of our proposed method, which regularizes the detection model with predictive information. As shown in Table~\ref{tab:plugnplay}, applying our approach to both existing models significantly improves the overall multi-camera 3D object detection performance. Note that, in this experiment, scores are from our reproduction on nuScenes~\cite{nuscenes} validation dataset. Also, we use four past observation frames (i.e., $N=4$) with the current frame.

\myparagraph{Comparison with SOTA approaches.}
Further, in Table~\ref{tab:validation}, our model based on BEVDepth~\cite{bevdepth} shows comparable performance to the current state-of-the-art approaches, including BEVDet, BEVDet4D, STS, BEVDepth, BEVStereo, SOLOFusion, and VideoBEV. Note that we apply CBGS~\cite{cbgs} technique to all models. We observe a notable improvement in metrics evaluating translation (mATE), velocity (mAVE), and orientation (mAOE). We would emphasize that adding more observation frames generally improves the overall detection performance, while our method can be easily applicable to other approaches as well. It would be worth exploring experiments with other better baselines (when their codes are released) to evaluate further whether similar improvements are observed. We leave it as our future work. 

\myparagraph{Per-Category Analysis.}
In Table~\ref{tab:categorical}, we provide a per-category (vehicles, pedestrian, bicycle) detection performance comparison between BEVDepth and ours. Though there is a degradation in mAOE for bicycles, we observe performance improvements in all metrics, including mATE, mASE, mAOE, and mAVE. In particular, pedestrian and bicycle performance improvements are more noticeable, possibly due to the effect of learning temporal cues for their behaviors, which helps to detect small dynamic objects.

\myparagraph{Effect of Each Components.}
Table~\ref{tab:ablation} shows the results of various experiments conducted to analyze the module combinations used in our method. In this experiment, ResNet-50 is used as the backbone, image resolution is 256 x 704, and BEVDet4D~\cite{bevdet4d} is used as the base model with 4 frames. Model A represents the baseline performance of BEVDet4D. The results of B demonstrate that the utilization of predicted BEV feature provides more insightful information for understanding the current scene compared to concatenating previous features used in BEVDet4D.
Experiment C provides evidence that the Prediction Encoder is a crucial module for utilizing past information to predict the current scene effectively. Experiments D and E demonstrate the usefulness of the Temporal-Spatial encoder and Dense encoder within the Prediction Encoder. F concludes that all these components collectively contribute to enhancing the overall performance of the model. 

\begin{table}[t]    
    \caption{Ablation study on nuScenes validation set. \textit{Abbr.} C: Use of Concatenated BEV frames, F: Use of Fusion Module, S: Spatiotemporal BEV Encoder, M: Use of Multi-resolution Features.}
    \label{tab:ablation}
    \renewcommand{\arraystretch}{1.2}
    \resizebox{\linewidth}{!}{%
        \begin{tabular}{lcccccc}
        \toprule
        Model & C & F & S & M & NDS $\uparrow$ & mAP $\uparrow$ \\ \midrule
        Model A (BEVDet4D~\cite{bevdet4d}) & \ding{52}  &    &     &   & 0.454 & 0.330  \\\midrule
        Model B &    & \ding{52}  & \ding{52}    & \ding{52} & 0.475  & 0.357  \\
        Model C & \ding{52}  & \ding{52}  &    &    & 0.481 & 0.361 \\
        Model D & \ding{52}  & \ding{52}  & \ding{52}   &    & 0.484     &  0.366       \\
        Model E & \ding{52}  & \ding{52}  &     & \ding{52}   &  0.485   & 0.360     \\\midrule 
        Model F (Ours)    & \ding{52}  & \ding{52}  & \ding{52}   & \ding{52} & \textbf{0.489} & \textbf{0.374}  \\ 
        \bottomrule
    \end{tabular}}
\end{table}

\begin{figure*}[t]
    \begin{center}
        \includegraphics[width=\linewidth]{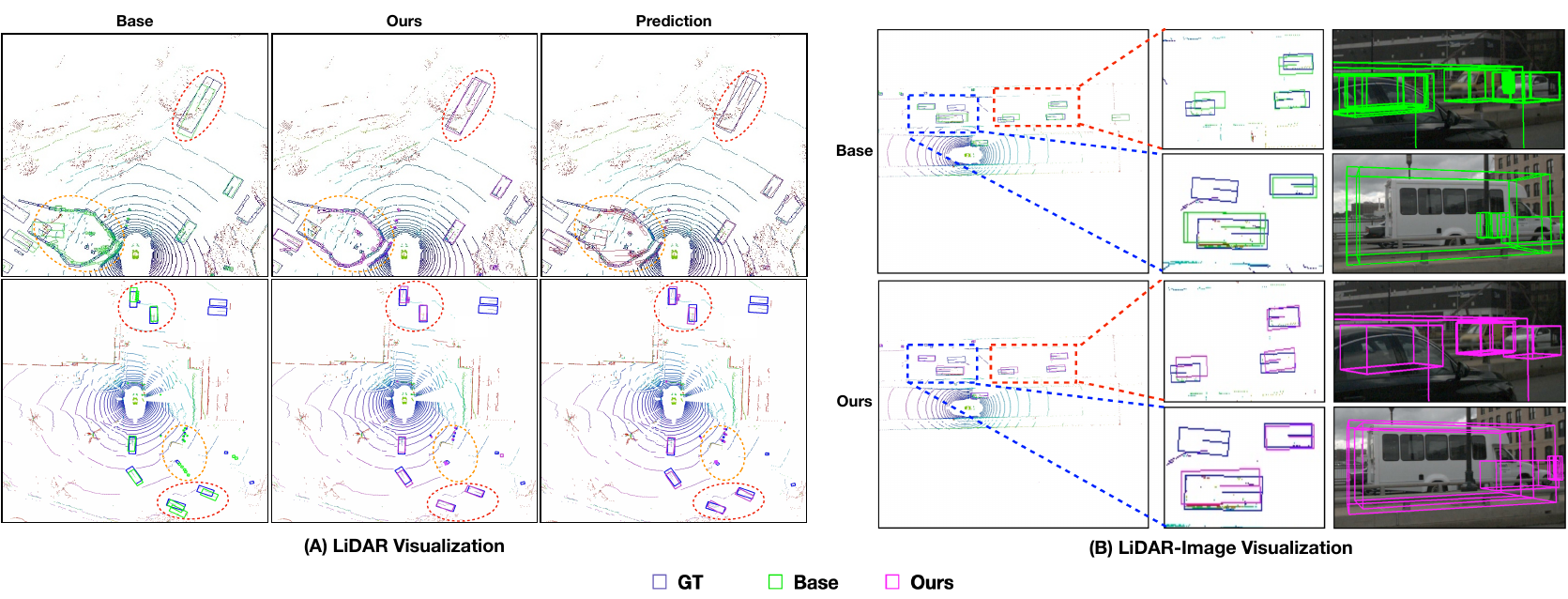}
    \end{center}
    \vspace{-.8em}
    \caption{Examples of detected objects (visualization on LiDAR and LiDAR-Image). (A) shows results from the base model, ours, and prediction (i.e., poses of predicted objects given only past observations). (B) provides a clearer view of the results from the base model and ours through visualization on LiDAR and image. The detected boxes in the Base model and Ours model are marked in green and purple, respectively. Ground truth boxes are marked in blue. }
    \label{fig:lidar_figure}
    \vspace{-1.em}
\end{figure*}

\begin{figure}[t]
    \begin{center}
        \includegraphics[width=\linewidth]{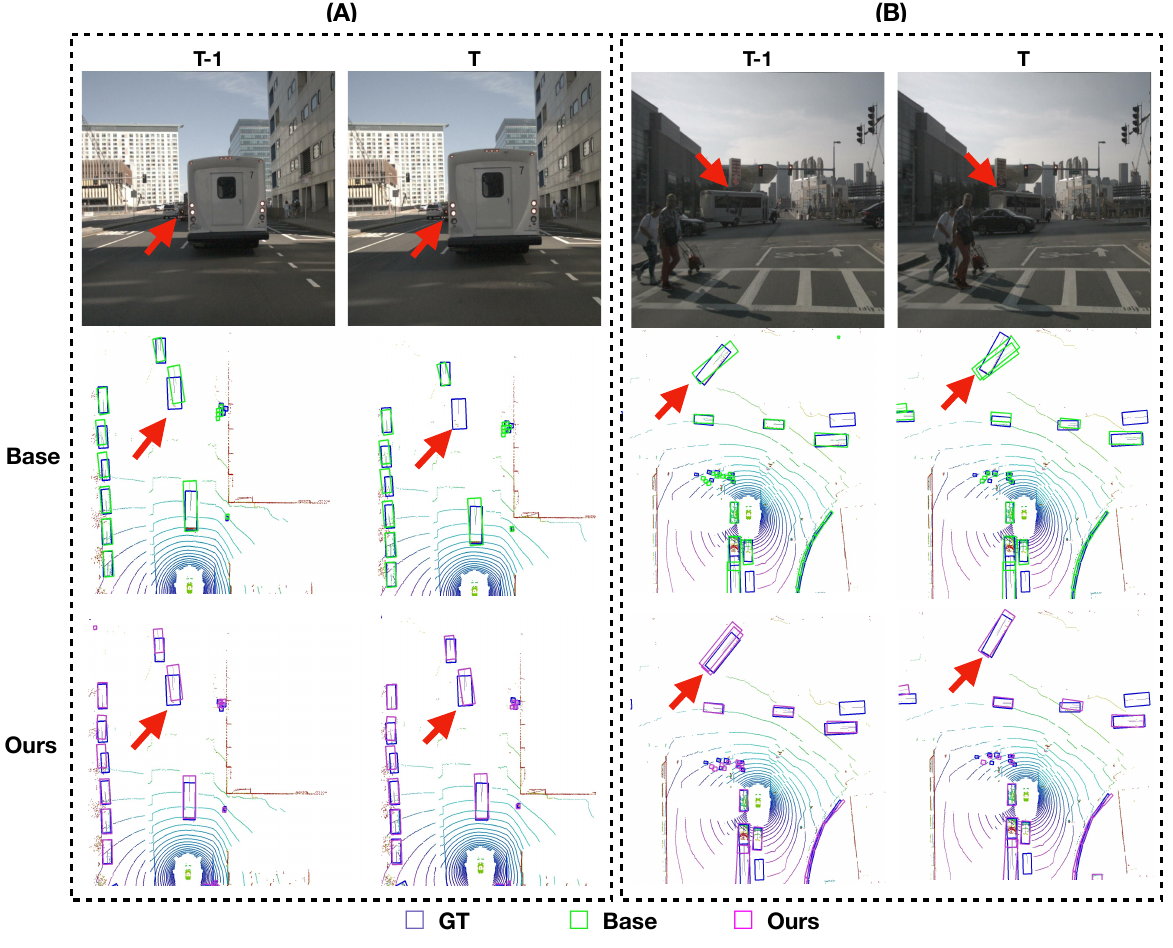}
    \end{center}
    \vspace{-1em}
    \caption{Visualization of two scenarios: one with occlusion occurring over time (A) and the other with an object making a left turn (B). The results from both scenarios demonstrate the contribution of predictions to object detection.}
    \label{fig:prediction}
    \vspace{-1.8em}
\end{figure}

\begin{table}[t]
    \caption{Performance comparison with variants of feature-level fusion methods. Data: nuScenes~\cite{nuscenes} validation set.}
    \label{tab:ablation_fusion}
    \renewcommand{\arraystretch}{1.3}
    \centering
    \resizebox{\linewidth}{!}{%
    \begin{tabular}{lcc}
    \toprule
    Model & NDS $\uparrow$ & mAP $\uparrow$ \\\midrule 
    A. Ours w/ Channel-wise Attention & 0.436 & 0.312   \\
    B. Ours w/ Concat followed by 1D ConvNet & 0.480  & 0.357 \\
    C. Ours w/ Concat followed by 2D ConvNet & 0.482  & 0.364 \\\midrule
    D. Ours (use of Deformable Attention)  & \textbf{0.489} & \textbf{0.374} \\
    \bottomrule
    \end{tabular}}
    \vspace{-2em}
\end{table}
\myparagraph{Comparison with other Fusion Methods.}
Table~\ref{tab:ablation_fusion} presents our experimental exploration of different fusion strategies for integrating Observed BEV features and Predicted BEV features within the Fusion Module. In this table, we conduct experiments exclusively on the validation dataset, without employing CBGS~\cite{cbgs}, and utilize four frames based on BEVDet4D~\cite{bevdet4d}.
(A) demonstrates the performance of the Channel-Wise Attention approach, where a learnable BEV feature is employed as the query, and attention is carried out by utilizing the features of the corresponding grid cells of $\mathbf{B}^{o}$ and $\mathbf{B}^{f}$ for key and value roles in order to extract information from each BEV grid cell. In (B) and (C), we employ concatenation to combine $\mathbf{B}^{o}\in \mathbb{R}^{H \times W \times D}$ and $\mathbf{B}^{f}\in \mathbb{R}^{H \times W \times D}$ to form a BEV feature with a shape of $\mathbb{R}^{H \times W \times 2D}$, followed by fusion using Conv1D and Conv2D methods, respectively, to extract BEV features with dimensions ${H \times W \times D}$. Lastly, in (D), we implement the Fusion Module approach as previously described. Our experimental findings confirm that the Deformable Attention-based method, where each feature $\mathbf{B}^{o}$ and $\mathbf{B}^{f}$ share offsets, is the most suitable fusion method.

\subsection{Qualitative Analysis}
In Fig.~\ref{fig:lidar_figure}, we visually present the results obtained from the nuScenes validation dataset. Predicted bounding boxes are represented in green, while the corresponding ground truth bounding boxes are shown in blue. 
In (A), the region within the red circle highlights an example where the Baseline model fails to accurately locate the object. However, by leveraging predictions that track object positions from previous features and combining them with the current feature, we demonstrate visually that more accurate detection can be achieved. The area delineated by the orange dashed line demonstrates how predictions can significantly reduce ghost boxes in regions where objects are densely concentrated.

(B) shows the results of the base model and our proposed model on objects with motion or occlusion. The visualization on the enlarged LiDAR and the corresponding image are shown together. The results show that our model is more accurately positioning the box.

As shown in Figure~\ref{fig:prediction}, the utilization of prediction information enables continuous detection even for occluded objects (see A) and improves the detection certainty for agents changing their direction (see B). This confirms that learning objects' motion further enhances the use of temporal cues, thereby enhancing overall detection accuracy. It demonstrates how predicting the current scene through the Temporal Context Extraction Module can be beneficial for detection.

\section{CONCLUSION}
In this paper, we propose a novel model as a way to utilize historical information to predict the current scene and improve detection performance in the current scene using the predictive information. Our model is designed to focus on the points that guide in better understanding the present by utilizing the continuous and temporally correlated features in the input data of autonomous driving or robotics. Through experiments on the NuScenes dataset, we have demonstrated the effectiveness of using predicted BEV features in multi-view 3D object detection. Our approach has shown performance improvements over the base model in terms of mAP, NDS metrics, as well as in translation, scale, orientation, and velocity. Furthermore, our model can be easily applied to 3D detection models that utilize images from previous timesteps in a plug-and-play manner. Future research could explore methods for refining predicted BEV features to leverage higher-quality prediction information.

\section*{ACKNOWLEDGMENT}
\small
This work was supported by Autonomous Driving Center, Hyundai Motor Company R\&D Division.

\bibliographystyle{IEEEtran}
\bibliography{ref}

\end{document}